\documentclass[10pt,twocolumn,letterpaper]{article}

\usepackage[pagenumbers]{cvpr} % To force page numbers, e.g. for an arXiv version

\usepackage{graphicx}
\usepackage{amsmath}
\usepackage{amssymb}
\usepackage{booktabs}

\usepackage{enumitem}
\usepackage{bm}
\usepackage{blindtext}
\usepackage{caption}
\usepackage{times}
\usepackage{epsfig}
\usepackage{pifont}
\usepackage{microtype}
\usepackage{colortbl}
\usepackage{xcolor}

\usepackage[pagebackref,breaklinks,colorlinks]{hyperref}

\usepackage[capitalize]{cleveref}
\crefname{section}{Sec.}{Secs.}
\Crefname{section}{Section}{Sections}
\Crefname{table}{Table}{Tables}
\crefname{table}{Tab.}{Tabs.}

\def\cvprPaperID{7112} % *** Enter the CVPR Paper ID here
\def\confName{CVPR}
\def\confYear{2023}

\begin{document}
	
\title{Spatio-Temporal Pixel-Level Contrastive Learning-based Source-Free \\ Domain Adaptation for Video Semantic Segmentation}
	
\author{Shao-Yuan Lo$^{1}$\thanks{This work was mostly done when S.-Y. Lo was an intern at Amazon.} \hspace{0.1cm} Poojan Oza$^{2}$ \hspace{0.1cm} Sumanth Chennupati$^{2}$ \hspace{0.1cm} Alejandro Galindo$^{2}$ \hspace{0.1cm} Vishal M. Patel$^{1}$ \\
$^{1}$Johns Hopkins University \hspace{0.1cm} $^{2}$Amazon \\
{\tt\small \{sylo, vpatel36\}@jhu.edu \hspace{0.1cm} \{poojanku, sumchenn, pagh\}@amazon.com}
%\and
}
\maketitle

%%%%%%%%% ABSTRACT
\begin{abstract}
Unsupervised Domain Adaptation (UDA) of semantic segmentation transfers labeled source knowledge to an unlabeled target domain by relying on accessing both the source and target data. However, the access to source data is often restricted or infeasible in real-world scenarios. Under the source data restrictive circumstances, UDA is less practical. To address this, recent works have explored solutions under the Source-Free Domain Adaptation (SFDA) setup, which aims to adapt a source-trained model to the target domain without accessing source data. Still, existing SFDA approaches use only image-level information for adaptation, making them sub-optimal in video applications. This paper studies SFDA for Video Semantic Segmentation (VSS), where temporal information is leveraged to address video adaptation. Specifically, we propose Spatio-Temporal Pixel-Level (STPL) contrastive learning, a novel method that takes full advantage of spatio-temporal information to tackle the absence of source data better. STPL explicitly learns semantic correlations among pixels in the spatio-temporal space, providing strong self-supervision for adaptation to the unlabeled target domain. Extensive experiments show that STPL achieves state-of-the-art performance on VSS benchmarks compared to current UDA and SFDA approaches. Code is available at: \url{https://github.com/shaoyuanlo/STPL}
\end{abstract}

%%%%%%%%% BODY TEXT
\section{Introduction} \label{sec:intro}
The availability of large amounts of labeled data has made it possible for various deep networks to achieve remarkable performance on Image Semantic Segmentation (ISS) \cite{badrinarayanan2017segnet,chen2017deeplab,lo2019efficient}. However, these deep networks often generalize poorly on target data from a new unlabeled domain that is visually distinct from the source training data. Unsupervised Domain Adaptation (UDA) attempts to mitigate this domain shift problem by using both the labeled source data and unlabeled target data to train a model transferring the source knowledge to the target domain \cite{ganin2015unsupervised,guan2021domain,lo2022exploring,lo2022learning,shin2021unsupervised,tsai2018learning}. UDA is effective but relies on the assumption that both source and target data are available during adaptation. In real-world scenarios, the access to source data is often restricted (e.g., data privacy, commercial proprietary) or infeasible (e.g., data transmission efficiency, portability). Hence, under these source data restrictive circumstances, UDA approaches are less practical.

\begin{figure}[!t]
	\centering
	\includegraphics[width=0.44\textwidth]{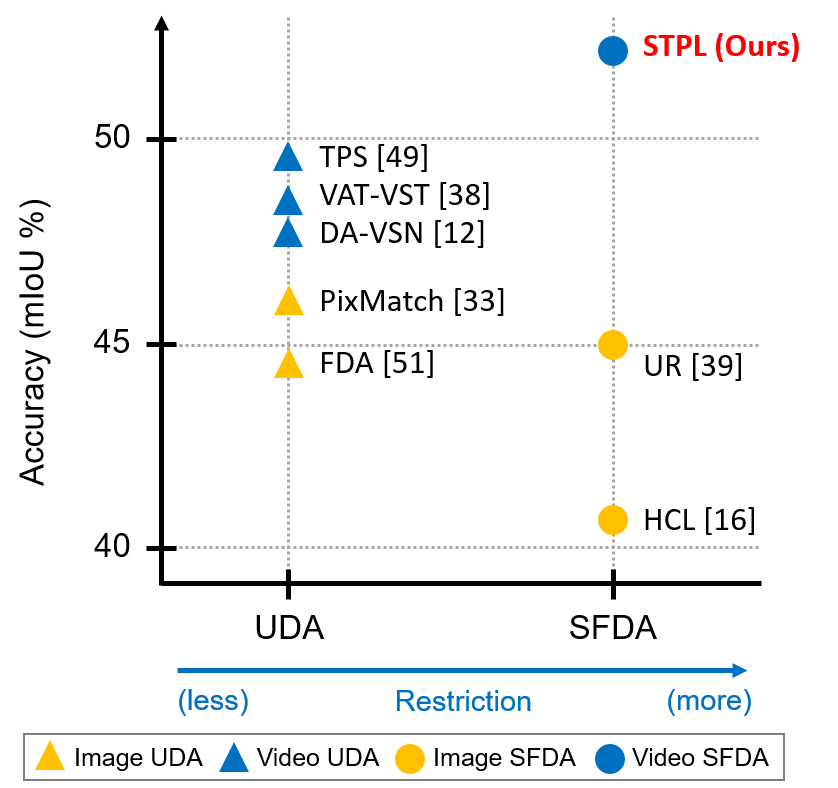}
	\caption{Comparison of VSS accuracy. Video-based UDA methods \cite{guan2021domain,shin2021unsupervised,xing2022domain} outperform image-based UDA methods \cite{melas2021pixmatch,yang2020fda}, showing the importance of video-based strategies for the VSS task. Image-based SFDA methods \cite{huang2021model,sivaprasad2021uncertainty} perform lower than the UDA methods, which shows the difficulty of the more restricted SFDA setting. The proposed STPL, even with SFDA, achieves the best accuracy and locates at the top-right corner of the chart (i.e., more restriction, but higher accuracy).}
	\label{fig:cover}
	%\vspace{-1.5em}
\end{figure}

To deal with these issues, the Source-Free Domain Adaptation (SFDA) setup, also referred to as Unsupervised Model Adaptation (UMA), has been recently introduced in the literature  \cite{chidlovskii2016domain,li2020model,liang2020we,yeh2021sofa}. SFDA aims to use a source-trained model (i.e., a model trained on labeled source data) and adapt it to an unlabeled target domain without requiring access to the source data. More precisely, under the SFDA formulation, given a source-trained model and an unlabeled target dataset, the goal is to transfer the learned source knowledge to the target domain. In addition to alleviating data privacy or proprietary concerns, SFDA makes data transmission much more efficient. For example, a source-trained model ($\sim$ 0.1 - 1.0 GB) is usually much smaller than a source dataset ($\sim$ 10 - 100 GB). If one is adapting a model from a large-scale cloud center to a new edge device that has data with different domains, the source-trained model is far more portable and transmission-efficient than the source dataset.
%Moreover, SFDA is more training efficient, as it only has to process target data and avoids the potential instability of adversarial learning in UDA approaches \cite{guan2021domain,shin2021unsupervised}.

Under SFDA, label supervision is not available. Most SFDA studies adopt pseudo-supervision or self-supervision techniques to adapt the source-trained model to the target domain \cite{sivaprasad2021uncertainty,huang2021model}. However, they consider only image-level information for model adaptation. In many real-world semantic segmentation applications (autonomous driving, safety surveillance, etc.), we have to deal with temporal data such as streams of images or videos. Supervised approaches that use temporal information have been successful for Video Semantic Segmentation (VSS), which predicts pixel-level semantics for each video frame \cite{jain2019accel,kundu2016feature,liu2020efficient,wang2021survey}. Recently, video-based UDA strategies have also been developed and yielded better performance than image-based UDA on VSS \cite{guan2021domain,shin2021unsupervised,xing2022domain}. This motivates us to propose a novel SFDA method for VSS, leveraging temporal information to tackle the absence of source data better. In particular, we find that current image-based SFDA approaches suffer from sub-optimal performance when applied to VSS (see Figure~\ref{fig:cover}). To the best of our knowledge, this is the first work to explore video-based SFDA solutions.

In this paper, we propose a novel spatio-temporal SFDA method namely Spatio-Temporal Pixel-Level (STPL) Contrastive Learning (CL), which takes full advantage of both spatial and temporal information for adapting VSS models. STPL consists of two main stages. (1) Spatio-temporal feature extraction: First, given a target video sequence input, STPL fuses the RGB and optical flow modalities to extract spatio-temporal features from the video. Meanwhile, it performs cross-frame augmentation via randomized spatial transformations to generate an augmented video sequence, then extracts augmented spatio-temporal features. (2) Pixel-level contrastive learning: Next, STPL optimizes a pixel-level contrastive loss between the original and augmented spatio-temporal feature representations. This objective enforces representations to be compact for same-class pixels across both the spatial and temporal dimensions.

With these designs, STPL explicitly learns semantic correlations among pixels in the spatio-temporal space, providing strong self-supervision for adaptation to an unlabeled target domain. Furthermore, we demonstrate that STPL is a non-trivial unified spatio-temporal framework. Specifically, \textit{Spatial-only CL} and \textit{Temporal-only CL} are special cases of STPL, and STPL is better than a naïve combination of them. Extensive experiments demonstrate the superiority of STPL over various baselines, including the image-based SFDA as well as image- and video-based UDA approaches that rely on source data (see Figure~\ref{fig:cover}). The key contributions of this work are summarized as follows:
\begin{itemize}[topsep=0pt,noitemsep,leftmargin=*]
	\item We propose a novel SFDA method for VSS. To the best of our knowledge, this is the first work to explore video-based SFDA solutions.
	\item We propose a novel CL method, namely STPL, which explicitly learns semantic correlations among pixels in the spatio-temporal space, providing strong self-supervision for adaptation to an unlabeled target domain.
	\item We conduct extensive experiments and show that STPL provides a better solution compared to the existing image-based SFDA methods as well as image- and video-based UDA methods for the given problem formulation.
\end{itemize}

%-------------------------------------------------------------------------
\section{Related work}
\noindent \textbf{Video semantic segmentation.}
VSS predicts pixel-level semantics for each video frame \cite{gadde2017semantic,hu2020temporally,jain2019accel,kundu2016feature,li2021dynamic,liu2020efficient}, which has been considered a crucial task for video understanding \cite{wang2021survey}. VSS networks use temporal information, the inherent nature of videos, to pursue more accurate or faster segmentation. For example, FSO \cite{kundu2016feature} employs the dense conditional random field as post-processing to obtain temporally consistent segmentation. NetWarp \cite{gadde2017semantic} uses optical flow information to transfer intermediate feature maps of adjacent frames and gains better accuracy. ACCEL \cite{jain2019accel} integrates predictions of sequential frames via an adaptive fusion mechanism. TDNet \cite{hu2020temporally} extracts feature maps across different frames and merges them by an attention propagation module. ESVS \cite{liu2020efficient} considers the temporal correlation during training and achieves a higher inference speed. These works rely on large densely annotated training data and are sensitive to domain shifts.

\begin{figure*}[!t]
	\centering
	\includegraphics[width=0.98\textwidth]{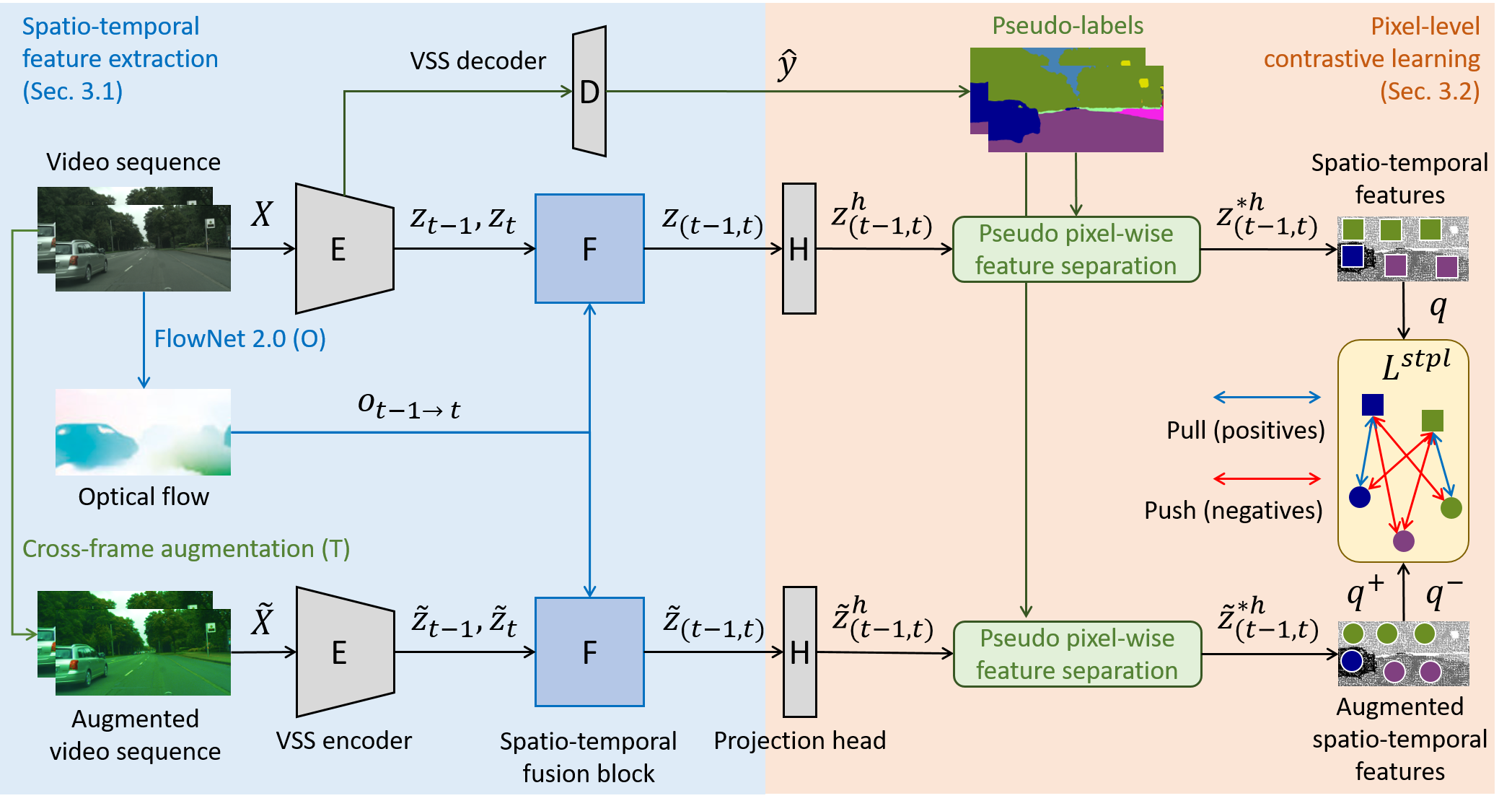}
	\caption{Overview of the proposed Spatio-Temporal Pixel-Level (STPL) contrastive learning framework.  STPL consists of two main stages. (1) Spatio-temporal feature extraction: First, STPL fuses the RGB and optical flow ($o_{t-1\to t}$) modalities to extract spatio-temporal features ($z_{(t-1,t)}$, $\tilde{z}_{(t-1,t)}$) from both the original and augmented video sequences ($X$, $\tilde{X}$). (2) Pixel-level contrastive learning: Next, after passing through a projection head and pseudo pixel-wise feature separation, STPL optimizes the pixel-level contrastive loss between the original and augmented spatio-temporal features ($\mathcal{L}^{stpl}$). For simplicity, this illustration considers a two-frame video sequence as the input.}
	\label{fig:framework}
	%\vspace{-1.5em}
\end{figure*}

\noindent \textbf{Unsupervised domain adaptation.}
UDA tackles domain shifts by aligning the representations of the two domains \cite{ganin2015unsupervised}. This framework has been widely investigated in ISS. Existing approaches can be categorized into two main streams: adversarial learning-based \cite{chang2019all,du2019ssf,tsai2018learning,vs2022meta,vs2021mega} and self-training-based \cite{huang2021rda,melas2021pixmatch,yang2020fda,zou2019confidence}. Recently, there are several works studying UDA for VSS \cite{guan2021domain,shin2021unsupervised,xing2022domain}. DA-VSN \cite{guan2021domain} presents temporal consistency regularization to minimize temporal discrepancy across different domains and video frames. VAT-VST \cite{shin2021unsupervised} extends both adversarial learning and self-training techniques to video adaptation. TPS \cite{xing2022domain} designs temporal pseudo supervision to adapt VSS models from the perspective of consistency training. These UDA approaches rely on labeled source data for adaptation, which is not practical in many real-world scenarios.

\noindent \textbf{Source-free domain adaptation.}
SFDA, a.k.a. UMA, aims to adapt a source-trained model to an unlabeled target domain without requiring access to the source data \cite{chidlovskii2016domain,li2020model,liang2020we,vs2023instance,yeh2021sofa}. It has been investigated for ISS in recent years \cite{huang2021model,kundu2021generalize,kundu2022balancing,liu2021source,sivaprasad2021uncertainty,stan2021unsupervised}. SFDA-SS \cite{liu2021source} develops a data-free knowledge distillation strategy for target domain adaptation. UR \cite{sivaprasad2021uncertainty} reduces the uncertainty of target data predictions. HCL \cite{huang2021model} presents the historical contrastive learning, which leverages the historical source hypothesis to compensate for the absence of source data. Edge/Feature-Mixup \cite{kundu2022balancing} generates mixup domain samples used for both source training and target adaptation. However, the need for modifying source training makes it inflexible, and it is expensive to be scaled to the video level. SFDA for videos is still relatively unexplored.

\noindent \textbf{Contrastive learning.}
CL has been a successful representation learning technique \cite{chen2020simple,he2020momentum,kim2021learning,khosla2020supervised,oord2018representation}. The key idea is to create positive and negative sample pairs, then learn disciminative feature representations by maximizing the embedding distance among positive pairs and minimizing that among negative pairs. Recent works \cite{alonso2021semi,wang2021exploring} further explore pixel-to-pixel contrast for the ISS task, but they need label supervision for training.

%-------------------------------------------------------------------------
\section{Proposed method}
An overview of the proposed STPL is illustrated in Figure~\ref{fig:framework}. STPL is implemented by two key designs: spatio-temporal feature extraction and pixel-level CL. This section first introduces the detailed designs. Then we demonstrate that STPL is a non-trivial unified spatio-temporal framework.

\subsection{Spatio-temporal feature extraction}  \label{sec:stfeat}
The input is an unlabeled target video sequence $X = \{x_1, x_2, ..., x_{t-1}, x_t\}$, where $x_t$ is the current frame. For simplicity, let us consider $X = \{x_{t-1}, x_t\}$, i.e., a video with a current frame and a previous frame. Given $X$, the VSS network's encoder $E$ extracts feature representations for each frame: $z_{t-1} = E(x_{t-1})$ and $z_t = E(x_t)$. In addition, we employ FlowNet 2.0 \cite{ilg2017flownet} denoted as $O$, a widely used optical flow estimator, to estimate the optical flow between the previous and the current frames as: $o_{t-1\to t} = O(x_{t-1}, x_t)$.

\begin{figure}[!t]
	\centering
	\includegraphics[width=0.46\textwidth]{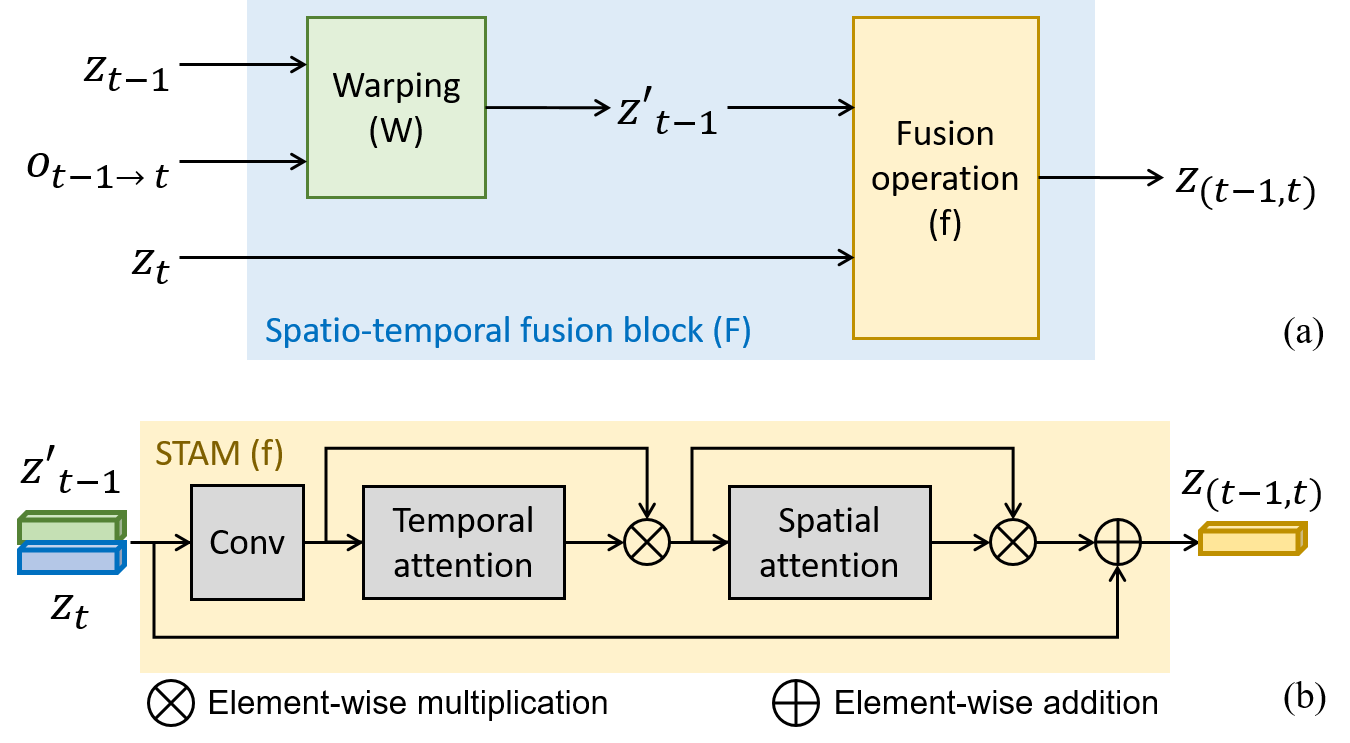}
	\caption{(a) The proposed spatio-temporal fusion block ($F$). (b) The proposed fusion operation ($f$): Spatio-Temporal Attention Module (STAM). STAM infers the attention of a spatio-temporal feature along the spatial and temporal dimensions separately, weighting important components in the spatio-temporal space. Details can be found in Appendix~\ref{sec:stam}. Our fusion block is also compatible with various fusion operations.}
	\label{fig:fusion}
\end{figure}

\noindent \textbf{Spatio-temporal fusion block.}
Next, we propose a spatio-temporal fusion block $F$ to extract spatio-temporal feature representations from the previous and the current features $z_{t-1}$ and $z_t$ (see Figure~\ref{fig:fusion} (a)). It adopts the estimated optical flow $o_{t-1\to t}$ to warp the previous feature $z_{t-1}$ to the propagated feature as: $z'_{t-1} = W(z_{t-1}; o_{t-1\to t})$, where $W$ denotes the warping operation. This feature propagation aligns the pixel correspondence between the previous and the current features, which is crucial for the dense prediction task. Then a fusion operation $f$ is used to fuse the cross-frame features into a spatio-temporal feature as: $z_{(t-1,t)} = f(z'_{t-1}, z_t)$.

The fusion operation integrates two input features into one output feature. It can be element-wise addition, concatenation, 1$\times$1 convolution layer, an attention module, or other variants. Inspired by \cite{woo2018cbam}, we design a Spatio-Temporal Attention Module (STAM) illustrated in Figure~\ref{fig:fusion} (b). STAM infers the attention of a spatio-temporal feature along the spatial and temporal dimensions separately, weighting important components in the spatio-temporal space. Details can be found in Appendix~\ref{sec:stam}.

\noindent \textbf{Cross-frame augmentation.}
Meanwhile, we perform cross-frame augmentation \cite{xing2022domain} that applies randomized spatial transformations $T$ on each input frame to generate an augmented video sequence: $\tilde{X} = T(X) = \{\tilde{x}_{t-1}, \tilde{x}_t\}$. Then we apply the same spatio-temporal feature extraction process on $\tilde{X}$ and extract the augmented spatio-temporal feature $\tilde{z}_{(t-1,t)}$. The augmentation $T$ contains randomized Gaussian blurring and color jittering transformations.

\subsection{Pixel-level contrastive learning}  \label{sec:plcl}
With the extracted original and augmented spatio-temporal features $z_{(t-1,t)}$ and $\tilde{z}_{(t-1,t)}$, we propose a new CL method to derive a semantically meaningful self-supervision. Typical CL schemes \cite{chen2020simple,khosla2020supervised} assume that an input contains only a single semantic category, and need a large batch size to offer sufficient positive/negative pairs for training. Nevertheless, in VSS, the input contains multiple instances, and a large batch size is computationally infeasible. Hence, we propose a method based on a pixel-level CL paradigm that leverages pixel-to-pixel contrast \cite{alonso2021semi,wang2021exploring}, and refer to our method as Spatio-Temporal Pixel-Level (STPL) CL.

\noindent \textbf{Pseudo pixel-wise feature separation.}
STPL aims to acquire pixel-level representations that are similar among the same-class pixel samples but distinct among different-class pixel samples. Since we do not have target domain labels, we use our VSS model's prediction for the input $X$ as pseudo-label $\hat{y}$. Subsequently, we use $\hat{y}$ to do pixel-wise feature separation. To maintain high-quality pseudo-labels, we set a hyperparameter of confident proportion $k$ to control the proportion of pixels preserved as pseudo-labels. More precisely, the confident pseudo-labels $\hat{y}^*$ are obtained by $\hat{y}^* = topk(\hat{y}; k) \subset\hat{y}$, where $topk$ is an operation that returns the $k$-proportion of the most confident predictions according to their probability scores.
% \cite{du2019ssf}. Note that our method is complementary to any pseudo-labeling techniques. More sophisticated versions \cite{shin2021unsupervised,zou2019confidence} also work for our STPL framework.

\noindent \textbf{Pixel-to-pixel contrastive loss.}
To perform CL, we first adopt a projection head $H$ to project our feature representations $z^h_{(t-1,t)} = H(z_{(t-1,t)})$ and $\tilde{z}^h_{(t-1,t)} = H(\tilde{z}_{(t-1,t)})$, similar to SimCLR \cite{chen2020simple}. According to the generated confident pseudo-labels $\hat{y}^*$, we denote the confident pixel representation sets in $z^h_{(t-1,t)}$ and $\tilde{z}^h_{(t-1,t)}$ as $z^{*h}_{(t-1,t)}\subset z^h_{(t-1,t)}$ and $\tilde{z}^{*h}_{(t-1,t)} \subset\tilde{z}^h_{(t-1,t)}$, respectively. Next, consider a query confident pixel representation $q\in z^{*h}_{(t-1,t)}$ (i.e., $q$ is a pixel representation in the feature $z^{*h}_{(t-1,t)}$) with a predicted pseudo-label $\hat{y}^*_q$, we define its positive pair set as:
% Positive pairs
\begin{equation}
	P_q \equiv \{q^+\in\tilde{z}^{*h}_{(t-1,t)} : \hat{y}^*_{q^+} = \hat{y}^*_q\},
\end{equation}
i.e., all the same-class pixels in the augmented feature $\tilde{z}^{*h}_{(t-1,t)}$. Then we define its negative pair set as:
% Negative pairs
\begin{equation}
	N_q \equiv \{q^-\in\tilde{z}^{*h}_{(t-1,t)} : \hat{y}^*_{q^-} \ne \hat{y}^*_q\},
\end{equation}
i.e., all the different-class pixels in $\tilde{z}^{*h}_{(t-1,t)}$. We follow SupCon \cite{khosla2020supervised} to develop a CL scheme with multiple positive pairs. The complete formulation of the proposed STPL contrastive loss is as follows:
% STPL loss
\begin{equation}  \label{eq:spatiotemporal}
	\mathcal{L}^{stpl}_q = \frac{-1}{|P_q|} \sum\limits_{q^+\in P_q} \log \frac{\exp(q\cdot q^+/\tau)}{\sum_{q^-\in N_q} \exp(q\cdot q^-/\tau)},
\end{equation}
where $\tau$ is a temperature parameter, and the $\cdot$ symbol denotes the inner product. Finally, the overall objective for the given video sequence input $X$ is defined as:
% STPL overall
\begin{equation}  \label{eq:overall}
	\mathcal{L}^{stpl} = \frac{1}{|z^{*h}_{(t-1,t)}|} \sum\nolimits_{q\in z^{*h}_{(t-1,t)}} \mathcal{L}^{stpl}_q.
\end{equation}
This objective enforces the pixel representations in the original spatio-temporal features to be similar to that of the same-class pixels in the augmented features, while being distinct from that of the different-class pixels. This explicitly learns semantic correlations among pixels in the spatio-temporal space and thus can achieve better class discriminability. The proposed STPL provides a strong self-supervision for video adaptation under the SFDA setup.

\subsection{STPL as a unified spatio-temporal framework}  \label{sec:unified}
We further demonstrate that STPL is a non-trivial unified spatio-temporal framework. Specifically, \textit{Spatial-only CL} and \textit{Temporal-only CL} are special cases of STPL. Moreover, we show that a naïve combination of them is sub-optimal compared to STPL.
%, i.e. \textit{the whole is greater than sum of its parts}.

\noindent \textbf{Spatial-only contrast.}
Let us turn off the fusion operation $F$ of the STPL framework with an identity operation. Then, let us allow only the current frame feature $z_t$, and similarly, only the augmented current frame feature $\tilde{z}_t$ to pass through the fusion block. After the projection head and confident filtering steps, the contrastive loss would be computed between $z^{*h}_t$ and $\tilde{z}^{*h}_t$ instead of the spatio-temporal features $z^{*h}_{(t-1,t)}$ and $\tilde{z}^{*h}_{(t-1,t)}$. That is, in Eq.~\eqref{eq:spatiotemporal} and Eq.~\eqref{eq:overall}, it becomes that $q\in z^{*h}_t$ and $\{q^+, q^-\}\in \tilde{z}^{*h}_t$. This computes contrast between only spatial variations and thus is a spatial-only special case of STPL. We denote this loss as $\mathcal{L}^{spa}$.

\noindent \textbf{Temporal-only contrast.}
Let us consider a duplicate copy of the input video as an augmentation (i.e., $\tilde{X}=X$). Next, let us turn off the fusion operation $F$ of STPL, allowing only the current frame feature $z_t$ and the augmented previous frame feature $\tilde{z}_{t-1}$ to pass through the fusion block. Here $\tilde{z}_{t-1} = z_{t-1}$ since $\tilde{X}=X$. Hence, after the projection head and confident filtering steps, the contrastive loss would be computed between $z^{*h}_t$ and $z^{*h}_{t-1}$. That is, in Eq.~\eqref{eq:spatiotemporal} and Eq.~\eqref{eq:overall}, it becomes that $q\in z^{*h}_t$ and $\{q^+, q^-\}\in z^{*h}_{t-1}$. This computes contrast between only temporal variations and thus is a temporal-only special case of STPL. We denote this loss as $\mathcal{L}^{tem}$.

\begin{figure}[!t]
	\centering
	\includegraphics[width=0.46\textwidth]{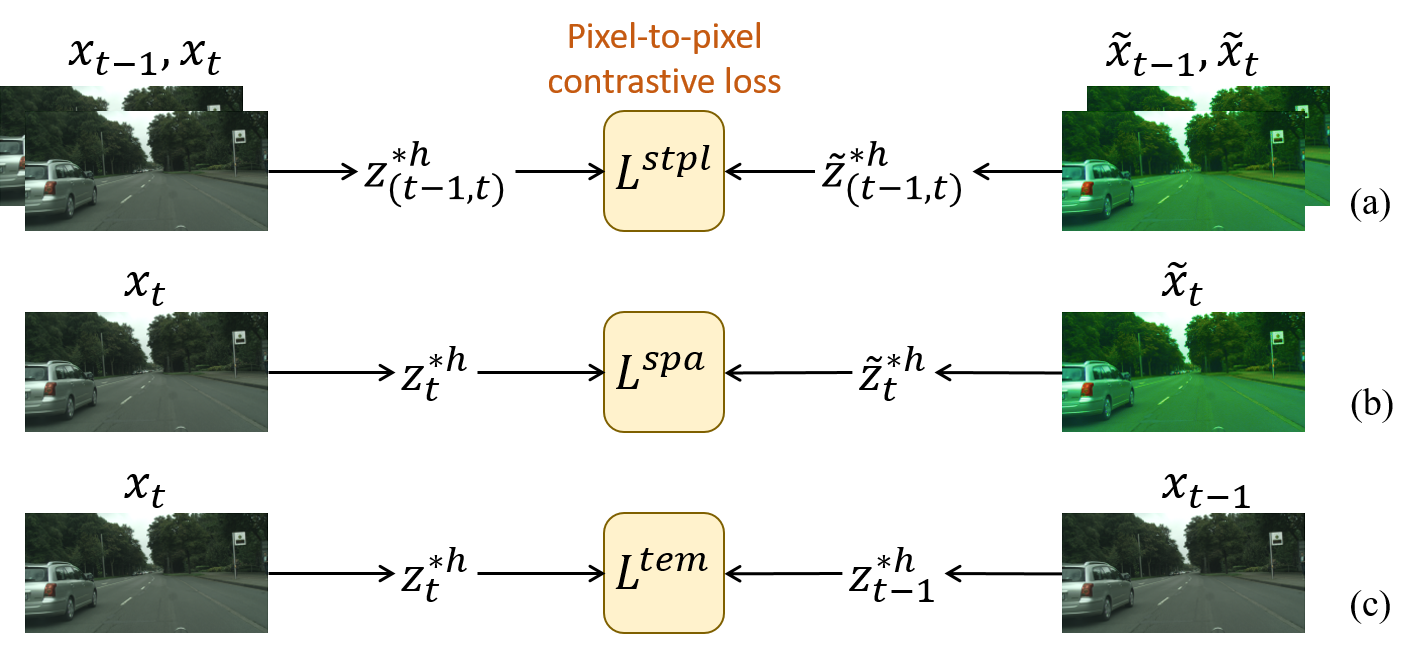}
	\caption{Illustration of (a) the proposed spatio-temporal contrast $\mathcal{L}^{stpl}$ (Eq.~\eqref{eq:spatiotemporal}, \eqref{eq:overall}), (b) spatial-only contrast $\mathcal{L}^{spa}$, and (c) temporal-only contrast $\mathcal{L}^{tem}$.}
	\label{fig:unified}
	%\vspace{-0.9em}
\end{figure}

\noindent \textbf{Naïve combination.}
To learn spatio-temporal contrast, a naïve way would be to combine the spatial-only and temporal-only contrastive losses together: $\mathcal{L}^{spa} + \mathcal{L}^{tem}$. Our experiments in Sec.~\ref{sec:ablation} show that the naïve combination is sub-optimal compared to STPL. This demonstrates that the proposed STPL is a non-trivial unified spatio-temporal framework. Figure~\ref{fig:unified} compares the proposed spatio-temporal contrast $\mathcal{L}^{stpl}$, spatial-only contrast $\mathcal{L}^{spa}$, and temporal-only contrast $\mathcal{L}^{tem}$.

%-------------------------------------------------------------------------
\section{Experiments}

\subsection{Experimental setup}
\noindent \textbf{Datasets.}
We evaluate our method on two widely used domain adaptive VSS benchmarks: VIPER \cite{richter2017playing} $\to$ Cityscapes-Seq \cite{cordts2016cityscapes} and SYNTHIA-Seq \cite{ros2016synthia} $\to$ Cityscapes-Seq. VIPER has 133,670 synthetic video frames with a resolution of 1080$\times$1920. SYNTHIA-Seq consists of 8,000 synthetic video frames with a resolution of 760$\times$1280. We consider VIPER and Synthia-Seq as source datasets to pre-train source models, respectively. Cityscapes-Seq is a realistic traffic scene dataset. It contains 2,975 training and 500 validation video sequences with a frame resolution of 1024$\times$2048. We use it as a target dataset. Following \cite{guan2021domain,xing2022domain}, we resize the frames of VIPER and Cityscapes-Seq to 760$\times$1280 and 512$\times$1024, respectively. For evaluations, the output predictions are interpolated to the original size.
%, where each sequence has 30 frames and the 20th frame is annotated.

\noindent \textbf{Implementation details.}
Following \cite{guan2021domain,xing2022domain}, we employ ACCEL \cite{jain2019accel} as our VSS network. It includes two segmentation branches, an optical flow estimation branch, and a prediction fusion layer. These branches consist of DeepLabv2 \cite{chen2017deeplab} architecture with ResNet-101 \cite{he2016deep} backbone, FlowNet \cite{dosovitskiy2015flownet}, and a 1$\times$1 convolution layer, respectively. All the adaptation models are trained by an SGD optimizer with an initial learning rate of $2.5e^{-6}$ and a momentum of 0.9 for 20k iterations. The learning rate decreases along the polynomial decay with a power of 0.9. We set the temperature $\tau=0.07$ and the confident proportion $k=0.7$. The mean Intersection-over-Union (mIoU) is used as the evaluation metric. Our experiments are implemented using PyTorch \cite{paszke2019pytorch}.

% VIPER
\renewcommand{\arraystretch}{0.9}
\setlength{\tabcolsep}{4.5pt}
\begin{table*}[!t]
	\scriptsize
	\begin{center}
		\caption{Quantitative comparisons (\%) with multiple types of domain adaptation baselines on VIPER \cite{richter2017playing} $\to$ Cityscapes-Seq \cite{cordts2016cityscapes}.}
		\label{table:viper}
		\begin{tabular}{r | r | r | rrrrrrrrrrrrrrr | r}
			\hline \noalign{\smallskip} \noalign{\smallskip}
			Method & Design & DA & road & side. & buil. & fence & light & sign & vege. & terr. & sky & pers. & car & truck & bus & mot. & bike & mIoU \\
			\noalign{\smallskip} \hline \noalign{\smallskip}
			Source-only & - & - & 56.7 & 18.7 & 78.7 & 6.0 & 22.0 & 15.6 & 81.6 & 18.3 & 80.4 & 59.9 & 66.3 & 4.5 & 16.8 & 20.4 & 10.3 & 37.1 \\
			\noalign{\smallskip} \hline \noalign{\smallskip}
			FDA \cite{yang2020fda} (CVPR'20) & Image & UDA & 70.3 & 27.7 & 81.3 & 17.6 & 25.8 & 20.0 & 83.7 & 31.3 & 82.9 & 57.1 & 72.2 & 22.4 & 49.0 & 17.2 & 7.5 & 44.4 \\
			PixMatch \cite{melas2021pixmatch} (CVPR'21) & Image & UDA & 79.4 & 26.1 & 84.6 & 16.6 & 28.7 & 23.0 & 85.0 & 30.1 & 83.7 & 58.6 & 75.8 & 34.2 & 45.7 & 16.6 & 12.4 & 46.7 \\
			RDA \cite{huang2021rda} (ICCV'21) & Image & UDA & 70.3 & 27.7 & 81.3 & 17.6 & 25.8 & 20.0 & 83.7 & 31.3 & 82.9 & 57.1 & 72.2 & 22.4 & 49.0 & 17.2 & 7.5 & 44.4 \\
			\noalign{\smallskip} \hline \noalign{\smallskip}
			UR \cite{sivaprasad2021uncertainty} (CVPR'21) & Image & SFDA & 84.2 & 20.1 & 80.1 & 11.5 & 30.7 & 31.1 & 82.8 & 22.1 & 69.2 & 59.5 & 81.0 & 4.9 & 52.7 & 36.6 & 8.7 & 45.0 \\
			HCL \cite{huang2021model} (NeurIPS'21) & Image & SFDA & 80.6 & 34.0 & 76.8 & 29.7 & 20.5 & 36.3 & 79.1 & 19.2 & 56.3 & 58.1 & 73.9 & 3.4 & 5.2 & 20.0 & 28.9 & 41.5 \\
			\noalign{\smallskip} \hline \noalign{\smallskip}
			DA-VSN \cite{guan2021domain} (ICCV'21) & Video & UDA & 86.8 & 36.7 & 83.5 & 22.9 & 30.2 & 27.7 & 83.6 & 26.7 & 80.3 & 60.0 & 79.1 & 20.3 & 47.2 & 21.2 & 11.4 & 47.8 \\
			VAT-VST \cite{shin2021unsupervised} (AAAI'22) & Video & UDA & 87.1 & 41.2 & 82.2 & 17.1 & 26.0 & 33.1 & 83.2 & 20.6 & 70.6 & 64.3 & 71.0 & 11.6 & 84.1 & 27.8 & 11.1 & 48.7 \\
			TPS \cite{xing2022domain} (ECCV'22) & Video & UDA & 82.4 & 36.9 & 79.5 & 9.0 & 26.3 & 29.4 & 78.5 & 28.2 & 81.8 & 61.2 & 80.2 & 39.8 & 40.3 & 28.5 & 31.7 & 48.9 \\
			\noalign{\smallskip} \hline \noalign{\smallskip}
			DA-VSN* \cite{guan2021domain} (ICCV'21) & Video & SFDA & 77.8 & 32.6 & 79.6 & 29.2 & 37.5 & 34.7 & 82.0 & 22.0 & 64.1 & 61.1 & 76.0 & 6.6 & 32.8 & 32.2 & 11.4 & 45.3 \\
			VAT-VST* \cite{shin2021unsupervised} (AAAI'22) & Video & SFDA & 48.2 & 20.4 & 78.1 & 28.8 & 33.1 & 33.6 & 81.1 & 20.0 & 56.1 & 58.3 & 74.7 & 8.6 & 73.5 & 29.7 & 9.6 & 43.6 \\
			TPS* \cite{xing2022domain} (ECCV'22) & Video & SFDA & 69.9 & 0.0 & 77.4 & 0.0 & 6.2 & 14.8 & 77.5 & 0.2 & 47.4 & 36.9 & 67.7 & 0.0 & 19.3 & 0.0 & 0.0 & 27.8 \\
			\noalign{\smallskip} \hline \noalign{\smallskip}
			STPL (Ours) & Video & SFDA & 83.1 & 38.9 & 81.9 & 48.7 & 32.7 & 37.3 & 84.4 & 23.1 & 64.4 & 62.0 & 82.1 & 20.0 & 76.4 & 40.4 & 12.8 & 52.5 \\
			\noalign{\smallskip} \hline \noalign{\smallskip}
			Oracle & - & - & 96.5 & 76.8 & 89.2 & 58.3 & 49.5 & 60.0 & 90.3 & 37.5 & 80.5 & 72.1 & 92.0 & 41.6 & 64.6 & 63.1 & 76.2 & 69.9 \\			
			\noalign{\smallskip} \hline
		\end{tabular}
	\end{center}
\end{table*}

% SYNTHIA
\setlength{\tabcolsep}{7.5pt}
\begin{table*}[!t]
	\scriptsize
	\begin{center}
		\caption{Quantitative comparisons (\%) with multiple types of domain adaptation baselines on SYNTHIA-Seq \cite{ros2016synthia} $\to$ Cityscapes-Seq \cite{cordts2016cityscapes}.}
		\label{table:synthia}
		\begin{tabular}{r | r | r | rrrrrrrrrrr | r}
			\hline \noalign{\smallskip} \noalign{\smallskip}
			Method & Design & DA & road & side. & buil. & pole & light & sign & vege. & sky & pers. & rider & car & mIoU \\
			\noalign{\smallskip} \hline \noalign{\smallskip}
			Source-only & - & - & 56.3 & 26.6 & 75.6 & 25.5 & 5.7 & 15.6 & 71.0 & 58.5 & 41.7 & 17.1 & 27.9 & 38.3 \\
			\noalign{\smallskip} \hline \noalign{\smallskip}
			FDA \cite{yang2020fda} (CVPR'20) & Image & UDA & 84.1 & 32.8 & 67.6 & 28.1 & 5.5 & 20.3 & 61.1 & 64.8 & 43.1 & 19.0 & 70.6 & 45.2 \\
			PixMatch \cite{melas2021pixmatch} (CVPR'21) & Image & UDA & 90.2 & 49.9 & 75.1 & 23.1 & 17.4 & 34.2 & 67.1 & 49.9 & 55.8 & 14.0 & 84.3 & 51.0 \\
			RDA \cite{huang2021rda} (ICCV'21) & Image & UDA & 84.7 & 26.4 & 73.9 & 23.8 & 7.1 & 18.6 & 66.7 & 68.0 & 48.6 & 9.3 & 68.8 & 45.1 \\
			\noalign{\smallskip} \hline \noalign{\smallskip}
			UR \cite{sivaprasad2021uncertainty} (CVPR'21) & Image & SFDA & 83.5 & 8.0 & 68.1 & 16.5 & 9.9 & 17.7 & 62.4 & 65.1 & 31.9 & 15.3 & 82.3 & 41.9 \\
			HCL \cite{huang2021model} (NeurIPS'21) & Image & SFDA & 79.0 & 44.7 & 78.9 & 25.4 & 12.9 & 36.6 & 75.2 & 63.0 & 49.0 & 19.5 & 50.1 & 48.6 \\
			\noalign{\smallskip} \hline \noalign{\smallskip}
			DA-VSN \cite{guan2021domain} (ICCV'21) & Video & UDA & 89.4 & 31.0 & 77.4 & 26.1 & 9.1 & 20.4 & 75.4 & 74.6 & 42.9 & 16.1 & 82.4 & 49.5 \\
			VAT-VST \cite{shin2021unsupervised} (AAAI'22) & Video & UDA & 82.8 & 26.5 & 78.3 & 23.7 & 12.8 & 20.0 & 78.4 & 64.5 & 45.5 & 16.0 & 69.6 & 47.1 \\
			TPS \cite{xing2022domain} (ECCV'22) & Video & UDA & 91.2 & 53.7 & 74.9 & 24.6 & 17.9 & 39.3 & 68.1 & 59.7 & 57.2 & 20.3 & 84.5 & 53.8 \\
			\noalign{\smallskip} \hline \noalign{\smallskip}
			DA-VSN* \cite{guan2021domain} (ICCV'21) & Video & SFDA & 81.0 & 37.9 & 68.4 & 23.7 & 14.0 & 27.5 & 69.8 & 71.3 & 46.4 & 18.7 & 80.2 & 49.0 \\
			VAT-VST* \cite{shin2021unsupervised} (AAAI'22) & Video & SFDA & 84.8 & 28.6 & 72.4 & 25.6 & 17.1 & 32.9 & 64.5 & 56.9 & 50.7 & 21.9 & 83.4 & 49.0 \\
			TPS* \cite{xing2022domain} (ECCV'22) & Video & SFDA & 62.6 & 0.0 & 69.2 & 0.2 & 0.8 & 14.4 & 56.6 & 10.4 & 4.2 & 0.2 & 24.5 & 22.1 \\
			\noalign{\smallskip} \hline \noalign{\smallskip}
			STPL (Ours) & Video & SFDA & 87.6 & 42.5 & 74.6 & 27.7 & 18.5 & 35.9 & 69.0 & 55.5 & 54.5 & 17.5 & 85.9 & 51.8 \\
			\noalign{\smallskip} \hline \noalign{\smallskip}
			Oracle & - & - & 96.4 & 78.1 & 89.1 & 43.6 & 42.3 & 64.9 & 90.3 & 84.4 & 66.8 & 50.7 & 92.7 & 72.7 \\
			\noalign{\smallskip} \hline
		\end{tabular}
	\end{center}
\end{table*}

\begin{figure*}[!t]
	\centering
	\includegraphics[width=0.98\textwidth]{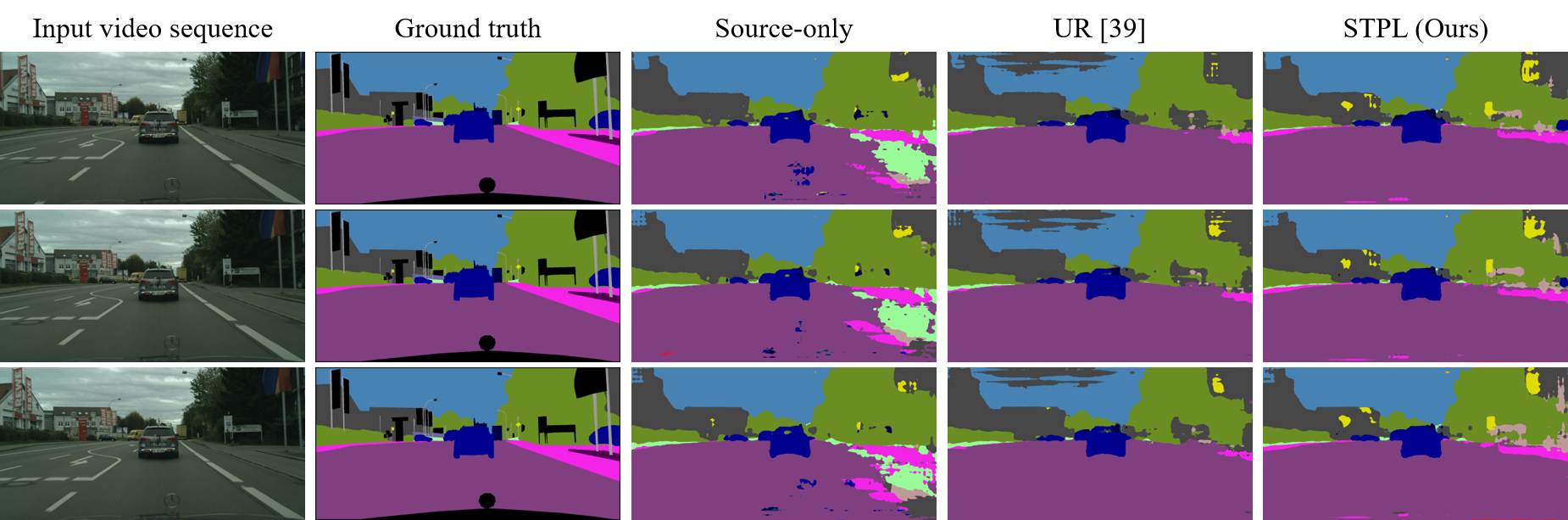}
	\caption{Qualitative results on VIPER \cite{richter2017playing} $\to$ Cityscapes-Seq \cite{cordts2016cityscapes}. The source-only model produces noisy and inconsistent predictions on the road and sidewalk. UR \cite{sivaprasad2021uncertainty}, an image-based SFDA method, suffers from inaccurate predictions on the sky and sidewalk. In contrast, the proposed STPL obtains more accurate segmentation results with high temporal consistency across the video sequence.}
	\label{fig:qualitative}
\end{figure*}

\subsection{Main results}
\noindent \textbf{Baselines.}
Since the proposed STPL is the first SFDA method for VSS, we compare it with multiple related domain adaptation state-of-the-art approaches described as follows. (1) \textit{Image-based UDA}: FDA \cite{yang2020fda}, PixMatch \cite{melas2021pixmatch} and RDA \cite{huang2021rda}; (2) \textit{Image-based SFDA}: UR \cite{sivaprasad2021uncertainty} and HCL \cite{huang2021model}; and (3) \textit{Video-based UDA}: DA-VSN \cite{guan2021domain}, VAT-VST \cite{shin2021unsupervised} and TPS \cite{xing2022domain}. The image-based approaches are applied to videos by using a VSS backbone (ACCEL in our experiments), following the practice of \cite{guan2021domain,xing2022domain}. Furthermore, to fairly assess our STPL, we create the SFDA versions of these video-based UDA approaches as our (4) \textit{Video-based SFDA} baselines. We remove all of their loss terms containing source data while keeping all the loss terms computed from only target data. We use the * symbol to denote these baselines. The results of the source-only and oracle (i.e., trained with target domain labels) models are also reported for reference. For fair comparisons, all four types of baselines use the same VSS backbone and training settings.

% \noindent \textbf{Results.}
\noindent \textbf{VIPER $\to$ Cityscapes-Seq.}
Table~\ref{table:viper} reports the evaluation results on the VIPER $\to$ Cityscapes-Seq adaptation benchmark. The proposed STPL outperforms all four types of baselines by decent margins, which is 15.1\% higher than the source-only model and 3.6\% higher than the best-performing competitor. In particular, its superiority over the image-based SFDA approaches indicates the benefits of a video-based solution and demonstrates the effectiveness of our spatio-temporal strategy for videos. We can also observe that the video-based UDA approaches suffer from performance degradation when applied to SFDA. Whereas, STPL achieves better performance even compared to their UDA results relying on source data.

\noindent \textbf{SYNTHIA-seq $\to$ Cityscapes-Seq.}
Table~\ref{table:synthia} provides the results on the SYNTHIA-Seq $\to$ Cityscapes-Seq benchmark. Similarly, our STPL is better than most baselines. Although TPS achieves the best accuracy under UDA, this requires accessing source data. Moreover, TPS*'s accuracy dramatically reduces to 22.1\% under SFDA, showing that it is not a proper solution when source data are unavailable. Overall, these results clearly demonstrate the superiority of STPL.

\noindent \textbf{Qualitative results.}
Figure~\ref{fig:qualitative} shows examples of qualitative results on VIPER \cite{richter2017playing} $\to$ Cityscapes-Seq \cite{cordts2016cityscapes}. The source-only model produces noisy and inconsistent predictions on the road and sidewalk, showing the domain shift effect. UR, an image-based SFDA method, suffers from inaccurate sky predictions and cannot detect the whole sidewalk. In contrast, the proposed STPL obtains more accurate segmentation results with high temporal consistency across the video sequence. This indicates the importance of a video-based strategy for the VSS task and demonstrates our method's effectiveness. The qualitative and quantitative results are consistent.

\subsection{Ablation analysis}  \label{sec:ablation}
\noindent \textbf{Objective functions.}
We conduct an ablation study to validate the effectiveness of our spatio-temporal objective for adaptation. We create several variants for comparison. \textit{Vanilla Self-training} simply computes the cross-entropy loss between predictions and pseudo-labels with a confident threshold. \textit{Duplicate CL} computes the pixel-level contrastive loss between two identical video frames, i.e., the loss described in Sec.~\ref{sec:plcl} but uses a duplicate copy as an augmentation and passes through the current frame features only. \textit{Temporal-only CL}, \textit{Spatial-only CL} and \textit{Naïve T+S CL} are described in Sec.~\ref{sec:unified}, whose objective functions are $\mathcal{L}^{tem}$, $\mathcal{L}^{spa}$ and $\mathcal{L}^{tem} +\mathcal{L}^{spa}$, respectively.

As can be seen in Table~\ref{table:objective}, the simple Duplicate CL achieves higher accuracy than Vanilla Self-training, showing the effectiveness of the pixel-level contrastive loss. Both Temporal-only CL and Spatial-only CL make an improvement over Duplicate CL, which indicates the importance of contrasting with variations. Naïve T+S CL, a naïve combination of the temporal-only and spatial-only contrastive losses, is slightly better than either single loss. The proposed spatio-temporal objective further outperforms Naïve T+S CL, showing that our design can learn more semantically meaningful context from the spatio-temporal space than simply adding the losses of two dimensions together. This demonstrates that our STPL is a non-trivial unified spatio-temporal framework for video adaptation.

\noindent \textbf{Fusion operations.}
As discussed in Sec.~\ref{sec:stfeat}, our STPL framework is compatible with various fusion operations used to extract spatio-temporal features. Here we consider and compare different fusion operations, such as element-wise addition, 1$\times$1 convolution layer, concatenation, and the proposed STAM module. In Table~\ref{table:fusion}, we can observe that STAM achieves the best performance, showing its effectiveness. On the other hand, adopting any fusion operation can outperform all the baselines in Table~\ref{table:viper} and variants in Table~\ref{table:objective}. This demonstrates that STPL maintains superior performance regardless of the choice of fusion operations.

% Ablation 1
\renewcommand{\arraystretch}{1.0}
\setlength{\tabcolsep}{14pt}
\begin{table}[t!]
	%\scriptsize
	\begin{center}
		\caption{Ablation study of different objective functions on VIPER \cite{richter2017playing} $\to$ Cityscapes-Seq \cite{cordts2016cityscapes}.}
		\label{table:objective}
		\begin{tabular}{l | l}
			\hline \noalign{\smallskip} \noalign{\smallskip}
			Method / Objective function & mIoU \\
			\noalign{\smallskip} \hline \noalign{\smallskip}
			Source-only & 37.1 \\
			\noalign{\smallskip} \hline \noalign{\smallskip}
			Vanilla Self-training & 45.4 (+8.3) \\
			Duplicate CL & 45.7 (+8.6) \\
			Temporal-only CL ($\mathcal{L}^{tem}$) & 47.4 (+10.3) \\
			Spatial-only CL ($\mathcal{L}^{spa}$) & 51.1 (+14.0) \\
			Naïve T+S CL ($\mathcal{L}^{tem} +\mathcal{L}^{spa}$) & 51.4 (+14.3) \\
			\noalign{\smallskip} \hline \noalign{\smallskip}
			STPL (Ours; $\mathcal{L}^{stpl}$) & \textbf{52.5 (+15.4)} \\
			\noalign{\smallskip} \hline
		\end{tabular}
	\end{center}
\end{table}

% Ablation 2
\setlength{\tabcolsep}{14pt}
\begin{table}[t!]
	%\scriptsize
	\begin{center}
		\caption{Ablation study of different fusion operations $f$ on VIPER \cite{richter2017playing} $\to$ Cityscapes-Seq \cite{cordts2016cityscapes}.}
		\label{table:fusion}
		\begin{tabular}{l | r}
			\hline \noalign{\smallskip} \noalign{\smallskip}
			Fusion operation & mIoU \\
			\noalign{\smallskip} \hline \noalign{\smallskip}
			Element-wise addition & 51.4 \\
			1$\times$1 convolution layer & 51.8 \\
			Concatenation & 52.3 \\
			STAM & \textbf{52.5} \\
			\noalign{\smallskip} \hline
		\end{tabular}
	\end{center}
	\vspace*{-\baselineskip}
\end{table}

\begin{figure*}[!t]
	\centering
	\includegraphics[width=0.98\textwidth]{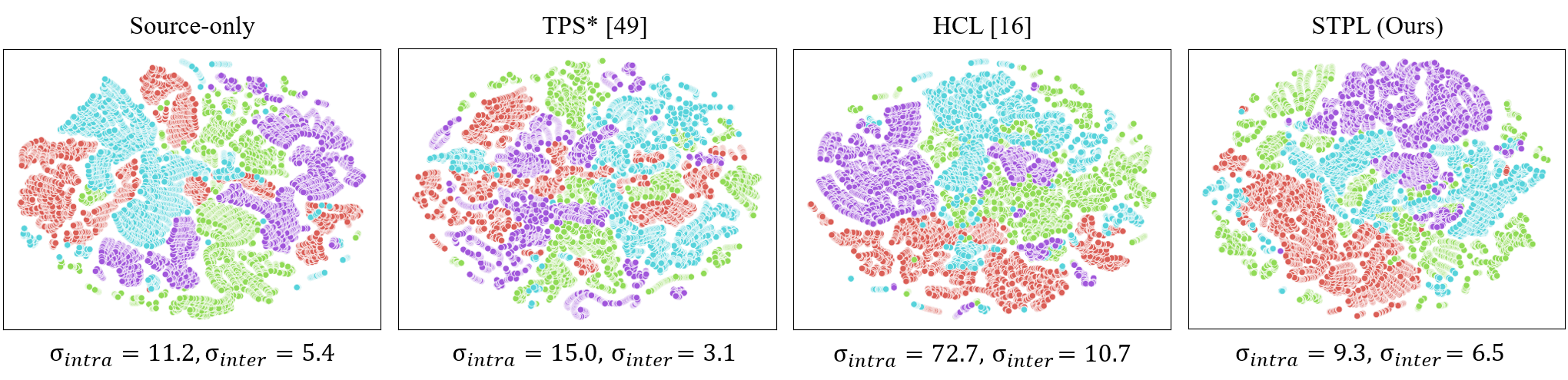}
	\caption{The t-SNE visualization \cite{van2008visualizing} of the feature space learned for VIPER \cite{richter2017playing} $\to$ Cityscapes-Seq \cite{cordts2016cityscapes}, where each point in the scatter plots stands for a pixel representation. Four classes (road, traffic light, car, and bicycle) are sampled to visualize. $\sigma_{intra}$ is the intra-class variance (lower is better) and $\sigma_{inter}$ is the inter-class variance (higher is better) of the feature space. All the methods are evaluated on the same selected video samples. In comparison, the proposed STPL learns the most discriminative feature space, which is reflected by the lowest $\sigma_{intra}$ and the high $\sigma_{inter}$.}
	\label{fig:tsne}
\end{figure*}

\begin{figure}[!t]
	\centering
	\includegraphics[width=0.39\textwidth]{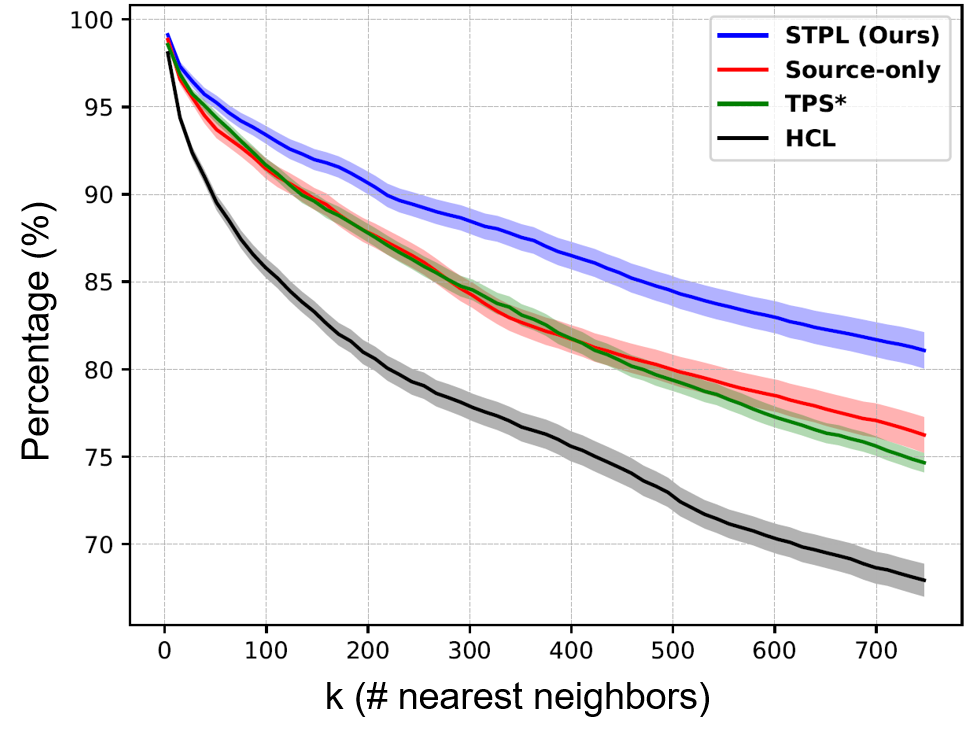}
	\caption{The percentage of same-class pixel representations among the $k$-nearest neighbors in the feature space. STPL achieves higher percentage for every $k$ value, showing that STPL learns a more discriminative and semantically consistent feature space.}
	\label{fig:knn}
\end{figure}

\noindent \textbf{Feature visualization.}
Figure~\ref{fig:tsne} provides the t-SNE visualization \cite{van2008visualizing} of the feature space learned for the VIPER $\to$ Cityscapes-Seq benchmark. For simplicity, we sample four classes (road, traffic light, car, and bicycle) to visualize. Each point in the scatter plots stands for a pixel representation. We compute the intra-class variance $\sigma_{intra}$ (lower is better) and inter-class variance $\sigma_{inter}$ (higher is better) of the feature space to provide a quantitative measurement. As can be seen, TPS*, which is originally designed for UDA, has a less discriminative feature space under the SFDA setup. It obtains a higher $\sigma_{intra}$ and a lower $\sigma_{inter}$ than the source-trained model. HCL, an image-based SFDA approach, acquires a higher $\sigma_{inter}$, but its $\sigma_{intra}$ is much higher. In comparison, the proposed STPL learns the most discriminative feature space. Unlike HCL, STPL leverages spatio-temporal information for video adaptation, and the benefit is clearly reflected by the lowest $\sigma_{intra}$ and the high $\sigma_{inter}$. This demonstrates STPL's ability to learn semantic correlations among pixels in the spatio-temporal space.

\noindent \textbf{Feature space neighborhood.}
This analysis inspects the neighborhood of the feature space learned by the proposed STPL, which quantitatively measures the discriminability of a feature space \cite{yang2021generalized}. We randomly select several video samples and extract the features at the pixel level. For an unbiased analysis, 500 pixel representations are considered for each semantic class to create a feature analysis set. Next, we query each representation in the set and retrieve the $k$-nearest neighbors of that representation. Among the retrieved $k$ nearest representations, we inspect the percentage of the same-class representations it contains.

Figure~\ref{fig:knn} reports the inspection results. For smaller $k$ values, all the methods have similar accuracy, which indicates that their feature spaces have semantically consistent neighbors for query pixel representations. Interestingly, when we increase the $k$ values to retrieve more neighbors, the accuracy differences between the proposed STPL and the other approaches significantly enlarge. In other words, the accuracy of STPL drops much slower than the rest. We can see that for any given $k$ values, STPL has more semantically consistent representations in the neighborhood. This analysis shows that our method effectively learns a discriminative feature space, thereby resulting in better performance.

%-------------------------------------------------------------------------
\section{Conclusion}
In this paper, we propose STPL, a novel SFDA method for VSS, which takes full advantage of spatio-temporal information to tackle the absence of source data better. STPL explicitly learns semantic correlations among pixels in the spatio-temporal space and provides strong self-supervision for video adaptation. To the best of our knowledge, this is the first work to explore video-based SFDA solutions. Moreover, we demonstrate that STPL is a non-trivial unified spatio-temporal framework. Extensive experiments show the superiority of STPL over various baselines, including the image-based SFDA as well as image- and video-based UDA approaches. Further insights into the proposed method are also provided by our comprehensive ablation analysis.

\noindent \textbf{Limitations.}
Similar to all the existing SFDA methods, STPL assumes that the source-trained model has learned source knowledge well. A sub-optimal source-trained model would affect adaptation performance. Such limitation of SFDA is an interesting direction for future investigations.

\noindent \textbf{Potential negative social impact.}
The proposed method may make attackers easier to adapt pre-trained open-source models to malicious uses. To avoid such risk, computer security or defense mechanisms could be incorporated.

%%%%%%%%% REFERENCES
{\small
	\bibliographystyle{ieee_fullname}
	\bibliography{my_cite}
}

%------------------------------------------------------------------------
\clearpage
\setcounter{section}{0}
\renewcommand\thesection{A\arabic{section}}

%\part*{Supplementary Materials}

\section{Details of the spatio-temporal fusion block} \label{sec:stam}
We design a fusion block specifically for spatio-temporal applications, namely Spatio-Temporal Attention
Module (STAM), as discussed in Sec.~\ref{sec:stfeat}. STAM is based on the attention mechanism inspired by \cite{woo2018cbam}. Consider the concatenation of the propagated previous feature $z'_{t-1}$ and the current feature $z_t$ as $z'_{(t-1,t)} \in \mathbb{R}^{T \times C \times H \times W}$, the STAM process can be written as:
% STAM
\begin{equation}
	\begin{split}
		z_{(t-1,t)} & = \{[A_{spa}[A_{tem}(z'_{(t-1,t)}) \otimes z'_{(t-1,t)}] \\ & \otimes [A_{tem}(z'_{(t-1,t)}) \otimes z'_{(t-1,t)}]\} \oplus z'_{(t-1,t)},
	\end{split}
\end{equation}
where $A_{tem}$ is temporal attension, $A_{spa}$ is spatial attension, $\otimes$ denotes element-wise multiplication, and $\oplus$ denotes element-wise addition.

\noindent \textbf{Temporal attention.}
The proposed temporal attention mechanism learns to choose informative temporal elements along each pixel's temporal dimension in the spatio-temporal space. The temporal attention $A_{tem} \in \mathbb{R}^{T \times 1 \times 1 \times 1}$ is performed as: 
% Temporal attention
\begin{equation}
	A_{tem}(z) = \sigma(FC(AvgPool(z)) + FC(MaxPool(z))),
\end{equation}
where $\sigma$ is the sigmoid function, and $FC$ denotes a fully connected layer.

\noindent \textbf{Spatial attention.}
The spatial attention mechanism chooses informative pixels along the spatial dimension in the spatio-temporal space. The spatial attention $A_{spa} \in \mathbb{R}^{1 \times 1 \times H \times W}$ is performed as: 
% Temporal attention
\begin{equation}
	A_{spa}(z) = \sigma(Conv(Concat[AvgPool(z), MaxPool(z)])),
\end{equation}
where $Concat$ denotes the concatenation operation, and $Conv$ denotes a convolutional layer.

\noindent \textbf{Remark.}
The main contribution of this paper is the STPL framework. In Table~\ref{table:fusion}, we can see that STPL can outperform all the existing methods even with the very simple Concatenation fusion, showing its flexibility. We propose STAM to show that STPL can further benefit from a more advanced fusion module.

\setlength{\tabcolsep}{10pt}
\begin{table}
	%\scriptsize
	\begin{center}
		\caption{Temporal consistency of different objective functions on VIPER \cite{richter2017playing} $\to$ Cityscapes-Seq \cite{cordts2016cityscapes}.}
		\label{table:temporal}
		\begin{tabular}{l | l}
			\hline \noalign{\smallskip} \noalign{\smallskip}
			Method / Objective function & Consistency (\%) \\
			\noalign{\smallskip} \hline \noalign{\smallskip}
			Source-only & 72.93 \\
			\noalign{\smallskip} \hline \noalign{\smallskip}
			Temporal-only CL ($\mathcal{L}^{tem}$) & 75.84 (+2.91) \\
			Spatial-only CL ($\mathcal{L}^{spa}$) & 77.68 (+4.75) \\
			Naïve T+S CL ($\mathcal{L}^{tem} +\mathcal{L}^{spa}$) & 80.91 (+7.89) \\
			\noalign{\smallskip} \hline \noalign{\smallskip}
			STPL (Ours; $\mathcal{L}^{stpl}$) & \textbf{82.14 (+9.21)} \\
			\noalign{\smallskip} \hline
		\end{tabular}
	\end{center}
\end{table}

\begin{figure}[!t]
	\centering
	\includegraphics[width=0.46\textwidth]{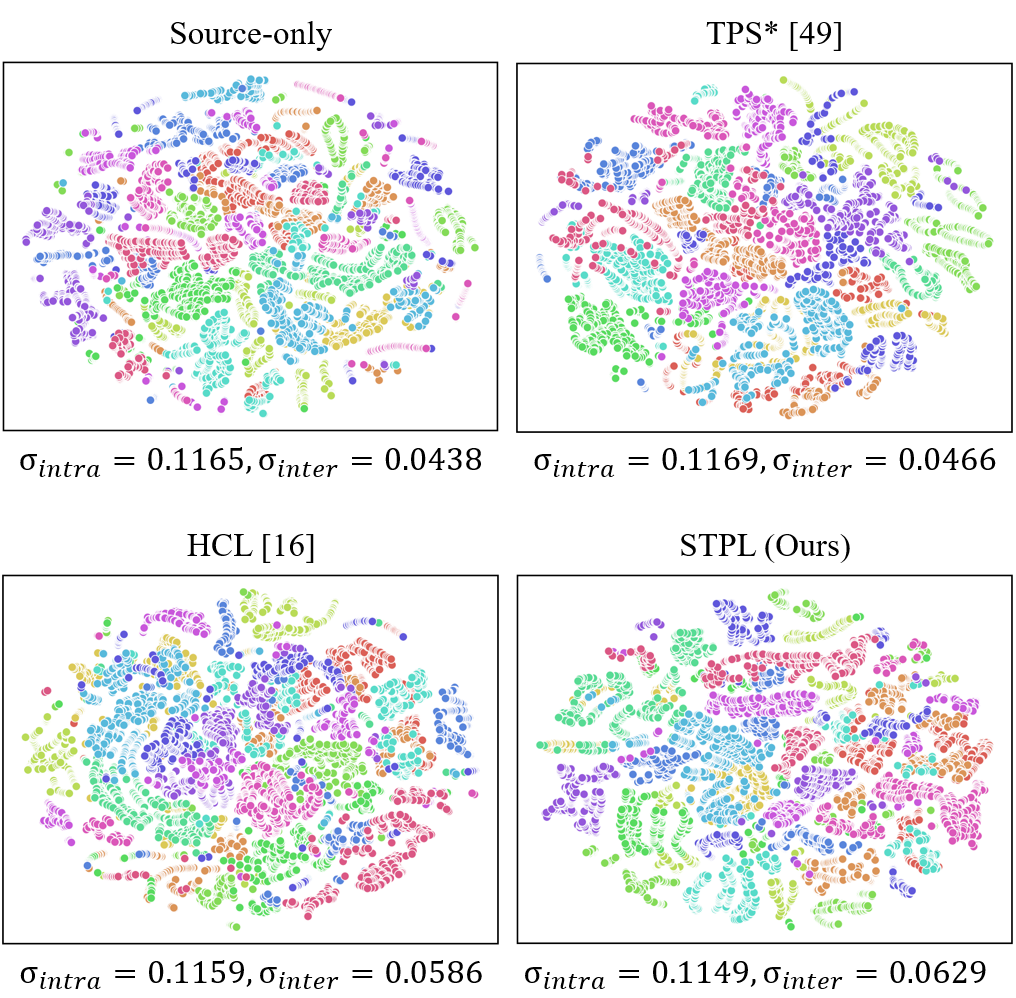}
	\caption{The t-SNE visualization \cite{van2008visualizing} of the feature space learned for VIPER \cite{richter2017playing} $\to$ Cityscapes-Seq \cite{cordts2016cityscapes}, where each point in the scatter plots stands for a pixel representation. All 15 classes are sampled to visualize. $\sigma_{intra}$ is the intra-class variance (lower is better) and $\sigma_{inter}$ is the inter-class variance (higher is better) of the feature space. All the methods are evaluated on the same selected video samples. In comparison, the proposed STPL learns the most discriminative feature space, which is reflected by the lowest $\sigma_{intra}$ and the highest $\sigma_{inter}$.}
	\label{fig:tsne2}
\end{figure}

\section{Temporal consistency}
We quantitatively compare the temporal consistency of different objective functions. The temporal consistency is derived from the overlap between the predicted segmentation maps of successive frames. We compute the percentage of the overlapping pixels. As shown in Table~\ref{table:temporal}, STPL performs the best, indicating that the proposed spatio-temporal method significantly improves temporal consistency. This quantitative result is consistent with the qualitative results shown in Figure~\ref{fig:qualitative}.

\section{More on feature visualization}
Figure~\ref{fig:tsne} provides the t-SNE visualization \cite{van2008visualizing} of the feature space learned for the VIPER $\to$ Cityscapes-Seq benchmark, where only four classes are sampled for simplicity. In this section, we visualize all 15 classes (see Figure~\ref{fig:tsne2}).
As can be seen, the proposed STPL learns the most discriminative feature space. It acquires the lowest $\sigma_{intra}$ and the highest $\sigma_{inter}$. This once again demonstrates STPL's ability to learn semantic correlations among pixels in the spatio-temporal space.

\end{document}